\documentclass[final,5p,times,twocolumn]{elsarticle}

\usepackage{amsmath,amssymb,amsfonts}
\usepackage{graphicx}
\usepackage{booktabs}
\usepackage{algorithm}
\usepackage{algpseudocode}
\usepackage{orcidlink}
\usepackage{hyperref}

\journal{Journal of Manufacturing Systems}

\begin{document}

\begin{frontmatter}

\title{Policy-Based Deep Reinforcement Learning Hyperheuristics for Job-Shop Scheduling Problems}

\author[inst1]{Sofiene Lassoued \orcidlink{0000-0001-7919-6939} *}
\author[inst2]{Asrat  Gobachew  \orcidlink{0000-0002-0118-6316}}
\author[inst2]{Stefan Lier      \orcidlink{0000-0002-3314-7610}}
\author[inst1]{Andreas Schwung  \orcidlink{0000-0001-8405-0977}}

\cortext[cor1]{Corresponding author: Sofiene Lassoued, email: lassoued.sofiene@fh-swf.de}

\affiliation[inst1]{organization={South Westphalia University of Applied Sciences}, 
                    department={Automation Technology and Learning Systems}, 
                    addressline={Lübecker Ring 2}, 
                    city={Soest}, 
                    postcode={59494}, 
                    state={North Rhine-Westphalia}, 
                    country={Germany}}

\affiliation[inst2]{organization={South Westphalia University of Applied Sciences}, 
                    department={Logistik und Supply Chain Management}, 
                    addressline={Lindenstr.53}, 
                    city={Meschede}, 
                    postcode={59872}, 
                    state={North Rhine-Westphalia}, 
                    country={Germany}}

\begin{abstract}

This paper proposes a policy-based deep reinforcement learning hyper-heuristic framework for solving the Job Shop Scheduling Problem. The hyper-heuristic agent learns to switch scheduling rules based on the system state dynamically. We extend the hyper-heuristic framework with two key mechanisms. First, action prefiltering restricts decision-making to feasible low-level actions, enabling low-level heuristics to be evaluated independently of environmental constraints and providing an unbiased assessment. Second, a commitment mechanism regulates the frequency of heuristic switching. We investigate the impact of different commitment strategies, from step-wise switching to full-episode commitment, on both training behavior and makespan. Additionally, we compare two action selection strategies at the policy level: deterministic greedy selection and stochastic sampling. Computational experiments on standard JSSP benchmarks demonstrate that the proposed approach outperforms traditional heuristics, metaheuristics, and recent neural network-based scheduling methods.

\end{abstract}


\end{frontmatter}

\section{Introduction}

The Shop Scheduling Problem (JSSP) is a fundamental and widely studied combinatorial optimization problem with significant practical relevance in many domains \cite{Zhang.2025}. Despite its widespread use, it remains computationally challenging due to its NP-complete nature \cite {Garey.1979}. Consequently, no algorithm can solve all JSSP instances optimally in polynomial time, despite numerous solution approaches ranging from exact to heuristics, metaheuristics, and learning-based approaches \cite{Chaudhry.2016}. 

Exact solutions, as Branch and Bound  \cite{Brucker.1994} Mixed-Integer Linear Programming\cite{Ku.2016}, Constraint Programming \cite{Beck.2011}, and dynamic programming \cite{Gromicho.2012}, suffer from the dimensionality curse with impractical solving time even for moderately sized problems. This led to a shift to near-optimal yet fast heuristic approaches. However, handcrafted heuristics are typically domain‑specific and require substantial problem expertise \cite{Kaban.2012}. For instance, a heuristic might excel in one scenario but fail to generalize to others.

Given these limitations, scientists experimented with nature-inspired metaheuristics to solve JSSPs, like Tabu search \cite{Ding.2015, Umam.2022}, genetic algorithms \cite{Deng.2015}, Simulated Annealing \cite{Wei.2018}, Ant Colony Optimization \cite{Huang.2008} and many other approaches. Still, metaheuristics require substantial algorithmic expertise, with parameter tuning often unrelated to the problem itself \cite{Burke.2009}, creating a domain barrier between optimization expertise and problem-domain expertise.

To address this domain barrier, the hyper-heuristic (HH) framework has been proposed. Widely used in scheduling problems \cite{Vela.2025}, high-level HH operate on low-level heuristics (LLHs) that construct solutions rather than search the solution space directly, thereby enabling improved generalization \cite{BOUAZZA.2023}. The high-level HH can take many forms, ranging from metaheuristics, primarily genetic programming (GP-HH), to more recently introduced deep reinforcement learning approaches (DRL-HH). Starting with the more widely used GP-HH, the authors of \cite{Burke.2009} examined the advantages and limitations of using GP as an HH. As an application, \cite{Liu.2017} employed GP-based HH to solve the Uncertain Capacitated Arc Routing Problem, while \cite{Guo.2024, Zhang.2022, Zhang.2021} applied genetic HH to tackle Dynamic Flexible Job-Shop Scheduling problems. 

Despite their advantages, GP-HH has several inherent limitations. First, each execution of GP can yield a different "best-of-run" heuristic, raising concerns about the approach's repeatability and predictability. Second, GP often requires unintuitive manual parameter tuning through trial-and-error to achieve competitive results \cite{Burke.2009}. In contrast, the knowledge-retention capability of Reinforcement Learning, implemented with a fixed seed, enables reproducible and consistent results. Moreover, unlike GP, RL hyperparameters are more theoretically grounded and easier to interpret. 

Deep reinforcement learning (DRL) has been successfully applied as a standalone optimization approach for the JSSP, achieving competitive performance with architectures ranging from graph neural networks \cite{Zhang.2020, Hameed.2023}, to attention- and transformer based models \cite{Chen.2023, Gebreyesus.2023}. However, despite their effectiveness, deep neural network–based methods alone generally suffer from limited explainability: unlike heuristic rules that are readily interpretable by domain experts, learned policy networks typically function as black boxes. This limitation motivates the integration of DRL with hyper-heuristics. On the one hand, HHs preserve domain interpretability by operating on low-level heuristics. On the other hand, DRL overcomes the limitations of traditional GP-HH by learning effective high-level heuristic-selection policies.

This paper builds on the potential of DRL-HH by addressing inherent challenges arising from combining the two hyper-heuristics and Deep Reinforcement learning frameworks, such as unbiased performance evaluation and credit assignment. The main contributions of this paper are summarized as follows: 

\begin{enumerate}
    \item We propose a novel policy-based deep reinforcement Learning Hyper-heuristics framework. The JSSP RL environment is modeled as a timed colored Petri net that explicitly captures scheduling dynamics and manufacturing constraints.

    \item We leverage the Petri net guard to ensure that only feasible low-level heuristics are considered in each state, allowing LLH performance to be evaluated independently of environmental constraints.
 
    \item We introduce \emph{commitment}. This temporal abstraction mechanism improves credit assignment in long-horizon scheduling by aggregating feedback over multiple consecutive timesteps, thereby enabling more stable learning.
    
    \item Our framework provides multi-level interpretability: action selections are inherently explainable by the selected dispatching rule, and real-time Petri net visualization further enables verification of the agent's decision-making process.

    \item The approach achieves state-aware adaptive dispatching that learns optimal heuristic selection across different scheduling contexts. Experimental results show our method outperforms static dispatching rules, metaheuristic algorithms, and end-to-end neural network approaches in a head-to-head comparison on the Taillard Benchmark. 


\end{enumerate}

\bigskip
This paper is structured as follows. Section~\ref{sec: related_work} reviews relevant literature and identifies the research gap addressed by this work. Section~\ref{sec: preliminaries} establishes the theoretical foundations, presenting the JSSP formulation, the timed colored Petri net representation, and the reinforcement learning framework based on the PPO algorithm. Section~\ref{sec: Hyper-Heuristic Framework} introduces our Hyper-heuristic methodology, detailing the JSSP environment design, the proposed approach, and its key advantages. Section~\ref{sec: Results and Discussion}  analyzes and discusses the results, examining the contribution of individual components of the proposed framework. Finally, Section~\ref{sec: Conclusion} summarizes the main findings and discusses directions for future research.

\section{Related Work}
\label{sec: related_work}
Hyper-heuristics (HH) aim at interchanging different solvers while solving a problem. The idea is to determine the best approach for solving a problem at its current state \cite{Sanchez.2020}. The HH framework separates the optimization process into two distinct domains: a control domain and a problem domain, separated by a domain barrier \cite{Burke.2013, Ozcan.2010}. This architectural separation offers several advantages. First, the problem domain is governed by low-level heuristics that are interpretable by domain experts, thereby preserving explainability. Second, this separation enables a modular design, allowing the high-level control strategy to be interchanged, for example, evolving from evolutionary methods to genetic algorithms or deep reinforcement learning, without modifying the problem-domain heuristic \cite{Pillay.2018}.

To examine the different subcategories of HH, we adopt the classification proposed in \cite{Burke.2019}, which organizes methods along two dimensions: the nature of the heuristic search space and the type of feedback mechanism. 

With respect to the search space, hyper-heuristics can be categorized into selection and generation approaches. Selection HH choose from a predefined set of low-level heuristics (LLHs) \cite{Drake.2020}. Generation HHs, by contrast, construct new heuristics by combining components of existing LLHs. In both categories, solutions can be built incrementally using constructive strategies or refined from a complete solution using perturbative strategies.

The second dimension concerns the feedback mechanism guiding the search. Three main categories can be distinguished: online learning, offline learning, and non-learning approaches \cite{Burke.2013}. In this paper, we focus on selection-based, learning-based hyper-heuristics, in which the optimizer is a deep reinforcement learning (DRL) agent. Inherited from the DRL framework, DRL-HH can be divided into value-based (VRL-HH) and policy-based (PRL-HH) approaches  \cite{Li.2024}.

Value-based methods span from traditional reinforcement learning–based hyper-heuristics (TRL-HH) to deep reinforcement learning (DRL) approaches. TRL-HH relies on techniques such as Q-table updates \cite{Choong.2018}, bandit-based updates \cite{Ferreira.2015}, and value estimation schemes \cite{Lamghari.2020} to evaluate and select the most suitable low-level heuristic (LLH) for a given state. In contrast, DRL-based methods integrate reinforcement learning with deep neural networks to handle larger and more complex state spaces. Examples include Deep Q-Networks (DQN) applied to routing problems \cite{Dantas.2021}, Double Deep Q-Networks (DDQN) for 2D packing problems \cite{Zhang.2022b}, and Dueling Double Deep Q-Networks (D3QN) for online packing problems \cite{Tu.2023}.

Owing to their simplicity, traditional reinforcement learning approaches dominate the literature, followed by Deep Q-learning. In contrast, policy-based approaches remain relatively scarce, despite their potential advantages.

On the policy-based hyper-heuristics side, the authors \cite{Udomkasemsub.2023} trained a Proximal Policy Optimization (PPO) hyper-heuristic agent to select generalized constructive low-level heuristics for combinatorial optimization problems, reporting improvements of up to 98\% over benchmark results. However, their constraint-handling strategy relies on post hoc penalization, in which solutions are first generated and then penalized for violating constraints, resulting in unnecessary computational effort spent evaluating invalid actions.

The authors of \cite{Kallestad.2023} also applied PPO-based hyper-heuristics to solve various combinatorial problems and compared their performance against Adaptive Large Neighborhood Search (ALNS). The authors introduced a "wasted action" mechanism to prevent the repeated selection of deterministic heuristics that do not change the system state, thereby avoiding an infinite loop. However, their choice of "insert" and "remove" operators as low-level heuristics limits the expressiveness of decisions relative to dispatching rules such as MTWR (Most total work remaining) or FIFO (First in, first out), thereby weakening one of hyper-heuristics's key advantages: interpretability for domain experts.

Applied to the trading domain, the authors \cite{Cui.2024} proposed a DRL-HH framework for multi-period portfolio optimization, reporting notable performance gains over state-of-the-art trading strategies and traditional DRL baselines. Although their approach employs rich state representations, the state-transition dynamics remain largely opaque; this is standard and acceptable in trading, where the environment is inherently stochastic and partially observable, and performance metrics such as return and risk are prioritized over interpretability. However, such opacity is less suitable for Job Shop Scheduling, where explainable system state evolution is of primary importance.

To address the  explainable system state evolution, we combined  in
our previous work \cite{Lassoued.2024, Lassoued.2025}, the modeling capabilities of timed colored Petri nets with the dynamic decision-making and knowledge retention of DRL. We also leveraged the Petri net's guard functions to mask invalid actions. Despite the improved explainability provided by the Petri net's graphical interface, the policy network itself remained a black box, leaving the decision-making process largely opaque. This motivates integrating our previous approach with a hyper-heuristic framework, which offers complementary benefits: hyper-heuristics LLH, enhance explainability. At the same time, dynamic action masking in Petri nets can improve the evaluation of low-level heuristics independently of the environment's constraints.

\bigskip
Research on DRL-based hyper-heuristics (DRL-HH) for the Job Shop Scheduling Problem faces several gaps and open challenges yet to be addressed. 

First, most existing HH approaches rely on tabular or score-based selection mechanisms, with limited use of actor–critic DRL models that can leverage rich state representations. Second, in many HH frameworks (e.g., VRL-HH), LLHs are penalized when proposed moves are rejected by move-acceptance strategies, even when rejections are caused by environmental constraints rather than poor heuristic choices, leading to biased performance signals; prefiltering infeasible actions is therefore necessary for fair LLH evaluation. Third, the literature lacks a systematic analysis of heuristic switching frequency, despite its strong impact on performance stability and credit assignment. Fourth, although HH frameworks offer inherent interpretability through domain-understandable LLHs, existing work does not explicitly capitalize on or enhance this explainability. Finally, direct and controlled comparisons between metaheuristics, DRL methods, and DRL-based HH on common JSSP benchmarks remain scarce.

To address these gaps, this paper employs a selection-based hyper-heuristic framework with a constructive methodology, using a policy-based deep reinforcement learning agent as an optimization algorithm. Data are collected through interactions with an environment that models the Job Shop Scheduling Problem using a colored timed Petri net. Additionally, the Petri net guard functions provide a natural pre-filter for move acceptance, thereby avoiding the evaluation of invalid actions.

\section{Preliminaries}
\label{sec: preliminaries}

In this section, we present the formulation and theoretical foundations of our approach. We begin by defining the Job Shop Scheduling Problem, including its objectives and constraints. We then introduce the mathematical formulation of the Petri net model and the reinforcement learning framework, followed by the theoretical background of Proximal Policy Optimization (PPO), which is used as the RL optimization algorithm.

\subsection{Job shop scheduling problem definition}

The Job Shop Scheduling Problem (JSSP) concerns the allocation of a set of \( n \) jobs \( \mathcal{J} = \{1, 2, \dots, n\} \) to a set of \( m \) machines \( \mathcal{M} = \{1, 2, \dots, m\} \). Each job \( j \in \mathcal{J} \) consists of an ordered sequence of operations \( O_{j1}, O_{j2}, \dots, O_{j\ell_j} \), where \( \ell_j \) denotes the number of operations in job \( j \). Each operation \( O_{jk} \) must be processed on a specific machine \( M_{jk} \in \mathcal{M} \) for a given processing time \( p_{jk} > 0 \). The goal is to determine feasible start times \( S_{jk} \) for each operation so as to minimize the makespan \( C_{\max} \), defined as the maximum completion time among all operations.

\begin{equation*}
\begin{aligned}
    &\text{(1)}\quad S_{j,k+1} \geq S_{jk} + p_{jk}, \quad \forall j \in \mathcal{J},\; k < \ell_j. \\
    &\text{(2)}\quad S_{jk} \geq S_{j'k'} + p_{j'k'} \;\text{or}\; S_{j'k'} \geq S_{jk} + p_{jk}, \\
    & \quad\quad \forall (j,k) \neq (j',k'),\; M_{jk} = M_{j'k'}. \\
    &\text{(3)}\quad C_{\max} \geq S_{jk} + p_{jk}, \quad\forall j \in \mathcal{J},\; k \leq \ell_j.\\
\end{aligned}
\end{equation*}

Here, Equation~(1) enforces the processing order of operations within each job, 
Equation~(2) ensures that no two operations overlap on the same machine, and 
Equation~(3) defines the makespan as the maximum completion time across all operations.

\subsection{Colored timed Petrinets}
\label{subsec: petrinet}

 Petri nets provide a formal graphical framework for modeling discrete-event systems characterized by concurrency, synchronization, and resource sharing. A Petri net is defined by the pair $(\mathcal{G}, \mu_0)$, where $\mathcal{G}$ is a bipartite graph of places $\mathcal{P}$ and transitions $\mathcal{T}$, and $\mu_0$ denotes the initial marking. Tokens represent resources or job states and move through the net when
transitions fire. For any node $n \in \mathcal{P} \cup \mathcal{T}$, let $\pi(n)$ and $\sigma(n)$ denote its input and output sets. A transition $t$ is enabled if every input place $p \in \pi(t)$ contains at least one token, and its firing updates the marking according to:

\begin{equation}
\Tilde{\mu}(p)=
    \begin{cases}
     \mu(p)-1 & p \in \pi(t),\\
     \mu(p)+1 & p \in \sigma(t),\\
     \mu(p)   & \text{otherwise}.
    \end{cases}
\end{equation}

Colored Petri nets (CPNs) extend this structure by associating data values (colors)
with tokens, enabling compact representations of systems with repeated but heterogeneous job structures. 
A CPN is defined as
\begin{equation}
\text{CPN} = (\mathcal{P}, \mathcal{T}, \mathcal{A}, \Sigma, \mathit{C},
\mathit{N}, \mathit{E}, \mathit{G}, \mathit{I}),
\end{equation}
where $\Sigma$ is the color set, $\mathit{C}$ assigns colors to nodes,
$\mathit{N}$ specifies arc directions, $\mathit{E}$ defines arc expressions,
$\mathit{G}$ encodes guard conditions, and $\mathit{I}$ provides the initial marking.  

Colored timed Petri nets (CTPNs) further extend CPNs by associating a transition with firing delays, such that a transition cannot fire until a token has spent the required sojourn time in its upstream place. This allows for modeling the processing times of a given operation. 

\subsection{Reinforcement learning framework }

Reinforcement Learning (RL) provides a framework for training agents to make sequential decisions through environmental interaction. An RL problem is formalized as a Markov Decision Process (MDP) defined by the tuple $(\mathcal{S}, \mathcal{A}, \mathcal{P}, \mathcal{R}, \gamma)$, where $\mathcal{S}$ represents the state space, $\mathcal{A}$ the action space, $\mathcal{P}(s' |s, a)$ the transition dynamics, $\mathcal{R}(s, a, s')$ the reward function, and $\gamma \in [0,1)$ the discount factor. The agent learns a policy $\pi(a|s)$ that maximizes the expected discounted return  :

\begin{equation}
\label{eq:return}
G_t = \sum_{k=0}^{\infty} \gamma^{k} R_{t+k+1}.
\end{equation}

RL algorithms can be categorized into value-based methods that estimate expected returns using value functions $V(s)$ or $Q(s,a)$, policy-based methods that directly optimize the policy parameters, and actor-critic methods that combine both approaches \cite{Richard.1998}. In this work, we adopt Proximal Policy Optimization (PPO) \cite{Schulman.2017} . PPO employs a clipped surrogate objective to constrain policy updates, ensuring stable training by preventing excessively large parameter changes. 

\subsection{Proximal Policy Optimization  algorithm }

The objective of the PPO agent  is to maximize the expected discounted return:

\begin{equation}
J(\theta) =
\mathbb{E}_{\tau \sim \pi_\theta}
\Big[
\sum_{t=0}^{\infty} \gamma^t r(s_t, a_t)
\Big],
\end{equation}

where $\tau$ denotes a trajectory generated by $\pi_\theta$. The policy gradient is estimated using the advantage-weighted log-probabilities:
\begin{equation}
\mathbb{E}_{s_t, a_t \sim \pi_\theta}
\Big[
\nabla_\theta \log \pi_\theta(a_t|s_t)\,
\hat{A}_t
\Big],
\end{equation}

where $\hat{A}_t$ is an  estimate of the advantage function. Advantage estimates are computed using a learned value function $V_\phi(s)$ and Generalized Advantage Estimation (GAE):
\begin{equation}
\hat{A}_t =
\sum_{l=0}^{\infty}
(\gamma \lambda)^l
\big(
r_{t+l} + \gamma V_\phi(s_{t+l+1}) - V_\phi(s_{t+l})
\big),
\end{equation}
with $\lambda \in [0,1]$ controlling the bias-variance tradeoff.

To stabilize learning, PPO introduces a clipped surrogate objective based on the probability ratio between the probability of an action in the old and new policy:
\begin{equation}
r_t(\theta) =
\frac{\pi_\theta(a_t|s_t)}{\pi_{\theta_{\text{old}}}(a_t|s_t)}.
\end{equation}

The clipped objective is defined as:
\begin{equation}
L^{\text{CLIP}}(\theta) =
\mathbb{E}_t
\Big[
\min\big(
r_t(\theta)\hat{A}_t,\;
\text{clip}(r_t(\theta), 1-\epsilon, 1+\epsilon)\hat{A}_t
\big)
\Big],
\end{equation}

which discourages excessively large policy updates. The full PPO loss includes a value function loss and an entropy bonus:
\begin{equation}
L(\theta, \phi) =
\mathbb{E}_t
\Big[
L^{\text{CLIP}}(\theta)
- c_1 (V_\phi(s_t) - R_t)^2
+ c_2 S[\pi_\theta](s_t)
\Big],
\end{equation}
where $R_t$ denotes the bootstrapped return used for value function training.
The parameters $(\theta, \phi)$ are optimized jointly using stochastic gradient descent.

\section{The PetriRL Hyper-Heuristic Framework}
\label{sec: Hyper-Heuristic Framework}

Since our hyperheuristic (HH) framework is built on reinforcement learning, it relies on the interaction between an agent and its environment. In this section, we first describe the environment, which captures the dynamics and constraints of the scheduling problem, and then introduce the HH agent that learns to select the most appropriate scheduling rule based on the observed system state. We conclude the section with the advantages of the proposed approach.

\subsection{The JSSP Environment}
\label{subsec: environment}

\begin{figure}[ht]
\centering
\includegraphics[width=1\linewidth]{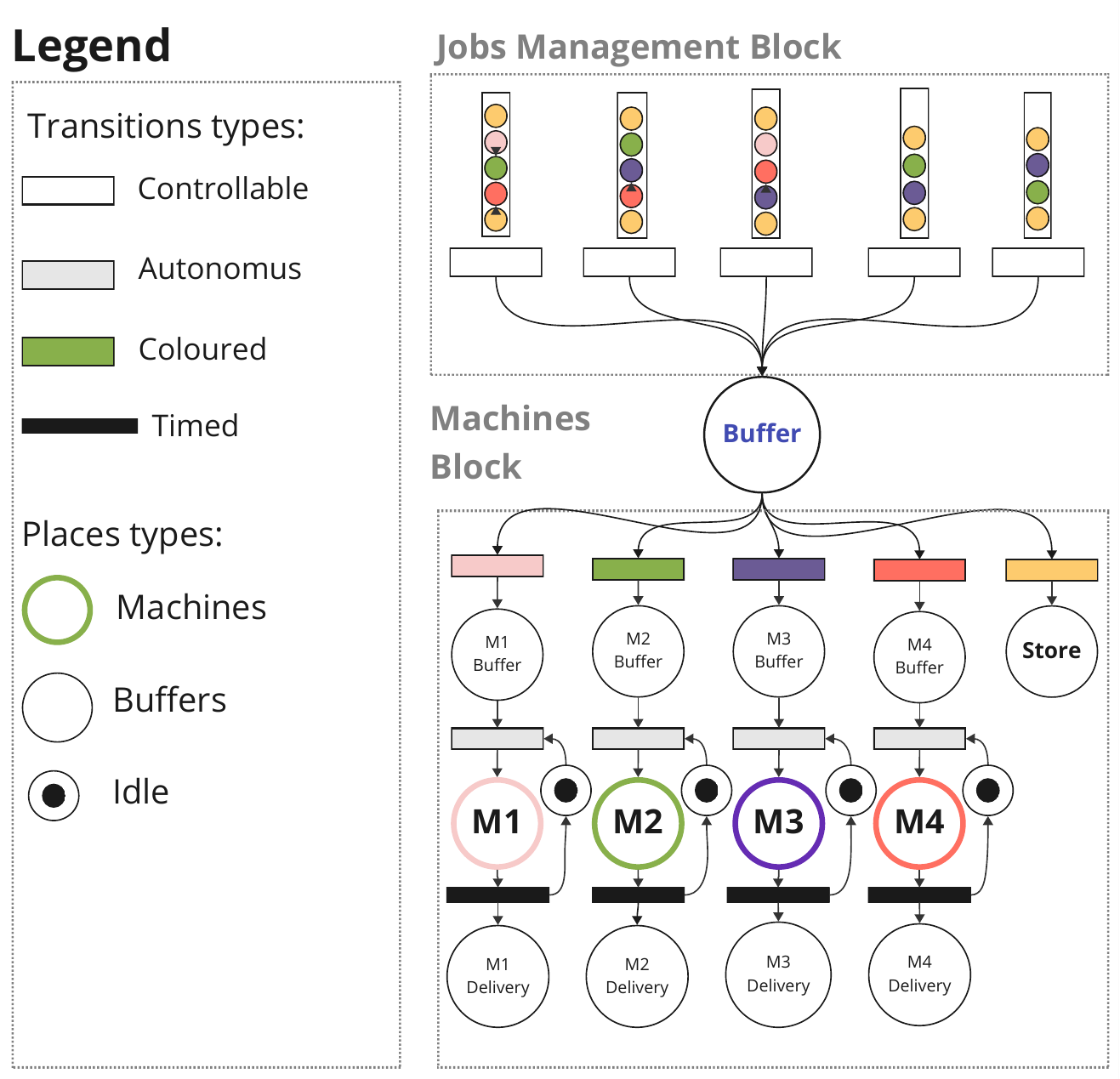}
\caption{Representation of a 5-job x 4-machine job shop scheduling problem modeled using a timed colored Petri net.}

\label{fig:env}
\end{figure}

The environment consists of a JSSP model and a reward function. We start with the JSSP model: following \cite{Lassoued.2024}, we employ a colored timed Petri net (CTPN) to simulate the dynamics of the scheduling problem while enforcing all constraints. In Figure~\ref{fig:env}, we depict a 5-job, 4-machine shop floor modeled using a CTPN.

Starting from the top of the figure, each job is represented as an ordered sequence of colored tokens. As introduced in Section~\ref{subsec: petrinet}, firing a transition moves a token from its parent place to a child place if the necessary conditions are met. Controllable transitions define the RL agent's action space, and the sequence of transition firings determines the job scheduling order. Once the RL agent selects the processing sequence, the remaining operations are managed automatically by the Petri net. Colored transitions route tokens to the appropriate machine buffers. If a machine is available, the token is processed. The processing time is captured by the sojourn time of the token at the machine place, which must elapse before the corresponding timed transition can fire.

The set of conditions, including the presence of a token in the parent place, the matching of the token's color, and the elapse of the sojourn time, defines the Petri net's guard function, which determines which transitions are enabled based on the current system state. This function is valuable not only as a constraint enforcer but also as an action-masking provider. By masking invalid actions, it significantly improves the efficiency and stability of the training process.

The key elements to define in any reinforcement learning environment are the observation space, action space, and reward function. As mentioned previously, the action space consists of the controllable transitions in the Petri net. The observation space comprises the distribution of tokens across places in the Petri net, also known as the \emph{marking}, augmented with additional temporal features such as elapsed time, remaining processing time, and machine availability. 

Concerning the reward function, we implement a terminal makespan penalty where the agent is trained to minimize the final makespan. The reward function is defined as:
\begin{equation} 
r_t = \begin{cases} 
-C_{\max}, & t = T,\\ 
0, & \text{otherwise}. 
\end{cases} 
\label{func: reward} 
\end{equation}
\noindent
Despite its simplicity, this reward function consistently provided the most robust results compared to more complex reward formulations. As Equation~(\ref{func: reward}) shows, this reward structure is sparse, providing feedback only at the end of each episode. We experimented with various reward shaping techniques to provide more frequent signals throughout the episode. However, most shaped rewards suffered from reward hacking, where the agent exploited shortcuts to maximize intermediate rewards while deteriorating the final makespan. The sparse terminal reward, though challenging for credit assignment, proved more reliable in guiding the agent toward genuinely optimal scheduling policies.

\subsection{PetriRL Hyper Heuristic overview}
\begin{figure}[ht]
\centering
\includegraphics[width=1\linewidth]{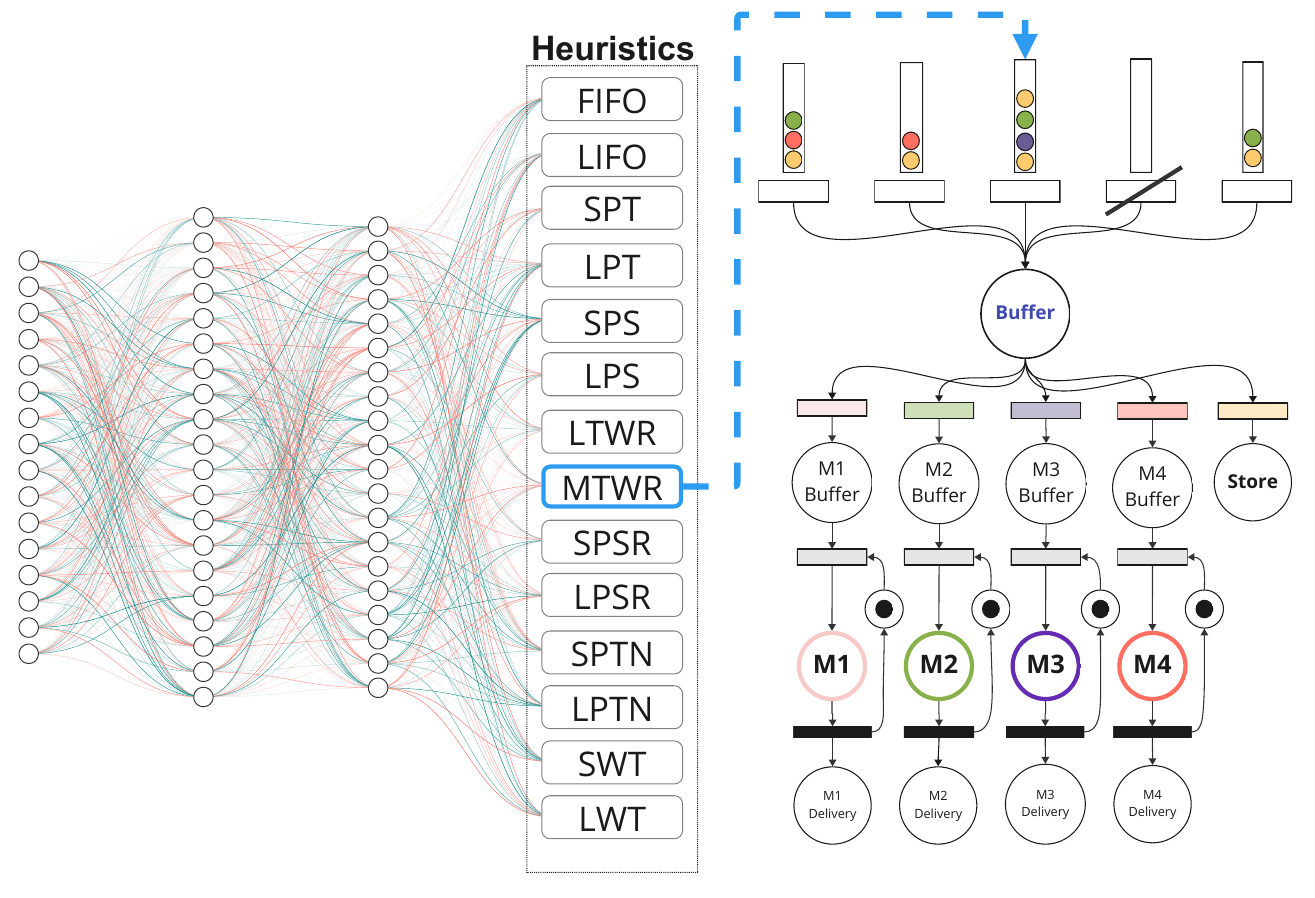}
\caption{PetriRL Hyper-heuristic framework: The policy network (left) selects an appropriate heuristic based on the current state. The chosen heuristic then decides the next action, which is executed in the colored timed Petri net environment modeling the JSSP (right).}
\label{fig: framework}
\end{figure}

In the proposed Hyperheuristic approach, instead of directly selecting job-machine assignments from a potentially large combinatorial action space, the agent operates at a higher level of abstraction by selecting dispatching rules from a predefined set of low-level heuristics. As depicted in Figure~\ref{fig: framework}, the agent's policy network observes the current system state and selects the most appropriate dispatching rule for the current scheduling context. The selected heuristic is then applied to determine which job-machine assignment should be executed, thereby updating the system state. 

In the proposed approaches, two action spaces are involved: the hyper-heuristics action space, composed of all the dispatching rules as low-level heuristics \(h_t \in \mathcal{H}\), and the environment-level action \(a_t \in \mathcal{A}\) produced by applying that heuristic rule to the current state \(s_t\) in the Petri net environment, for example fire transition T2 that allocates and operation from job 1 to machine 2. Let the set of low-level heuristics be :
\begin{equation}
\mathcal{H} = \{ h_1, h_2, \dots, h_{N} \}.
\end{equation}
Each heuristic \(h_k \in \mathcal{H}\) is a deterministic mapping
\begin{equation}
h_k : \mathcal{S} \to \mathcal{A}.
\end{equation}

The RL agent selects a heuristic according to the policy:

\begin{equation}
\pi_\theta(k \mid s_t) = P_\theta(h_k \mid s_t),
\end{equation}

and the chosen heuristic produces the actual dispatching action under the dynamic
masking induced by the Petri net guard function, this means the Petri net Per-filter the invalid action and provides the low-level heuristic only with possible actions to decide from :

\begin{equation}
a_t = h_k(s_t \mid \mathrm{mask}(s_t)).
\end{equation}

A commitment parameter is introduced: once a heuristic \(h_k\) is selected, it is applied for \(x\) consecutive steps. The pseudo code describing the approach is below:  

\begin{algorithm}[H]
\caption{PPO Hyperheuristic for JSSP}
\begin{algorithmic}[1]
\State \textbf{Input:} Environment $\mathcal{E}$, heuristics $\mathcal{H}$, policy $\pi_\theta$, commitment length $x$, episodes $N$
\State \textbf{Output:} Trained policy $\pi_\theta^*$

\For{episode $=1$ to $N$}
    \State Reset $\mathcal{E}$ and observe $s_0$
    \While{not done}
        \State Sample heuristic index $k \sim \pi_\theta(k \mid s_t)$
        \State Commit to $h_k$ for $x$ steps

        \For{$i = 0$ to $x-1$}
            \State Compute valid actions: $\mathcal{A}_{\text{valid}} \gets \{a : G(a)=1\}$
            \State Select $a_{t+i} \gets h_k(s_{t+i}, \mathcal{A}_{\text{valid}})$
            \State Execute $a_{t+i}$, observe $r_{t+i}$ and $s_{t+i+1}$
            \State $R_t \gets R_t + r_{t+i}$
        \EndFor

        \State Store $(s_t, k, R_t)$
        \State $t \gets t + x$
    \EndWhile
    \State Update $\pi_\theta$ using PPO
\EndFor

\State \Return $\pi_\theta^*$
\end{algorithmic}
\end{algorithm}


\subsection{Analysis of the approach's features}

The proposed approach offers several advantages, which we will evaluate in the results section. 
These include a fixed-size action space, improved credit assignment, adaptive state-dependent switching, enhanced interpretability, and the ability to learn mixtures of heuristics.

\subsubsection{Reduced and fixed Action Space size}

Without Hyper-heuristics, the RL action space consists of the controllable Petri net transitions, which represent decisions for job selection and machine allocation:
\begin{equation}
|\mathcal{A}| = (\text{jobs waiting}) \times (\text{machines}),
\end{equation}
Growing combinatorially with problem size. By switching to a hand-selected set of LLHs, the action space is drastically reduced.The dimension of the action space becomes :

\begin{equation}
|\mathcal{A}| = |\mathcal{H}| , \quad\mathcal{H}= \{ h_1, h_2, \dots, h_{N} \}.
\end{equation}

This not only drastically reduces the complexity of the learning problem but also creates a fixed action space that is independent of the number of jobs or machines. Consequently, the learned PPO policy is size-agnostic and can generalize across different problem sizes, while benefiting from improved stability and sample efficiency.

\subsubsection{Policy Search over Heuristics}
\label{subsec: Policy Search over Heuristics}

Since each dispatching heuristic $h_k$ can be represented as a deterministic HH policy that selects the same heuristic at every decision point, the HH policy class strictly subsumes the set of single-heuristic policies. Therefore, in principle, the optimal HH policy cannot perform worse than the best individual heuristic:
\begin{equation}
\max_{\pi \in \Pi} J(\pi) \;\ge\; \max_k J(h_k),
\end{equation}
where $\Pi$ denotes the set of all HH policies. 

In practice, whether a learned HH actually outperforms the best heuristic depends on the optimization algorithm, instance diversity, and problem dynamics.
Empirical studies suggest that adaptive selection often yields higher returns than fixed heuristics \cite{Cowling.2001}. We evaluate this behavior in our experiments and present the results in Section~\ref{sec: Results and Discussion}.

\subsubsection{State-Dependent Switching}
\label{subsec: State-Dependent Switching}

The authors of \cite{Kaban.2012} used simulation software to model and analyze 44 dispatching rules, consisting of 14 single rules and 30 hybrid rules across many performance criteria. They found that no single dispatching rule achieves optimal performance across all measured criteria. Nevertheless, the study highlighted key insights: SPT (Shortest Processing Time) excels at minimizing queue time, SPS (Shortest Process Sequence) performs well in reducing WIP (Work In Process), and LWT (Longest Waiting Time) is effective for minimizing queue length. Although MTWR (Most Total Work Remaining) emerged as the most consistently effective single rule overall, these findings demonstrate the potential for intelligent, state-dependent rule selection, where an adaptive agent can dynamically choose the most suitable rule based on the current system state, seamlessly switching from one dispatching rule to another as conditions change.

\subsubsection{Rule commitment and effect Credit Assignment}
In this section, we discuss the effect of introducing rule commitment on the credit assignment problem. In Section \ref{subsec: environment}, we established the motivation for using the makespan as a reward function. However, this creates a credit assignment challenge. Without commitment, the policy gradient becomes :

\begin{equation}
\nabla_\theta J \approx \sum_{t=0}^{T-1} \nabla_\theta \log \pi_\theta(a_t \mid s_t) \left(-\gamma^{T-t} C_{\max} - V_\phi(s_t)\right).
\end{equation}
\noindent
Each timestep receives an advantage signal diluted across all $T$ decisions, making it difficult to identify which heuristic choices actually contributed to the final makespan.

By committing to heuristic $h_k$ for $x$ consecutive steps, we reduce the effective decision horizon from $T$ to $M = T/x$. And each decision now collects the advantages from all x commitment steps. The policy gradient under commitment becomes:

\begin{equation}
\nabla_\theta J_{\text{commit}} \approx \sum_{m=0}^{M-1} \nabla_\theta \log \pi_\theta(h_k^m \mid s_{t_m}) \sum_{i=0}^{x-1} A_{t_m+i}.
\end{equation}
\noindent

Commitment introduces a form of temporal abstraction, transforming the decision process from per-event control to higher-level heuristic selection, conceptually related to temporally extended actions in the Options framework \cite{Sutton.1999}. The commitment mechanism can improve credit assignment through two complementary effects. First, aggregating advantages over $x$ consecutive steps provides a more robust evaluation signal for each heuristic choice. Instead of assessing a heuristic based on a single timestep's outcome, which may be noisy or unrepresentative, the agent evaluates it based on its cumulative performance over multiple timesteps. Second, reducing the number of sequential decisions from $T$ to $T/x$ simplifies the attribution problem. 

We empirically validate these benefits in Section \ref{sec: ablation}, comparing heuristic commitment versus per-step switching on identical JSSP instances.

\subsubsection{Interpretability}

In the framework, we leverage one of the biggest advantages of HH: using low-level, explainable heuristics that are often handcrafted by domain experts, without sacrificing performance. For instance, each action corresponds to a known scheduling rule $h_k$, thereby allowing the agent's behavior to be explained in a natural and interpretable manner. For example, in state $s_t$, the agent may select the MTWR rule with probability $\pi_\theta(k \mid s_t)$ = 80\%, indicating a preference for prioritizing jobs with the largest remaining workload under the current system conditions.

This interpretability is further enhanced by our specific use of HH within a {colored timed Petri net framework. The Petri net provides a real-time, human-understandable graphical representation of the system state previously discussed in section~\ref{sec: ablation}, enabling users to verify the agent's decisions visually. In particular, the user can directly observe that the selected action targets the job with the most remaining work, as indicated by the job place with the most significant number of tokens waiting in the queue.

\section{Results and Discussion}
\label{sec: Results and Discussion}

\subsection{Experimental Setup}

We evaluated our approach against a diverse set of methods from the literature for solving Job Shop Scheduling Problems. All methods were benchmarked on the widely used Taillard dataset~\cite{Taillard.1993} to enable a direct comparison of performance. The Taillard benchmark comprises eight groups of instances with varying sizes, ranging from $15 \text{ jobs} \times 15 \text{ machines}$ to $100 \text{ jobs} \times 20 \text{ machines}$. Within each size group, ten instances are provided, each differing in processing times, job sequences, and machine allocations .

The competing algorithms include learning-based neural network models such as the Disjunctive Graph Embedded Recurrent Decoding Transformer (DGERD) \cite{Chen.2023}, Gated-Attention Model (GAM) \cite{Gebreyesus.2023}, and Graph Isomorphism Network (GIN) \cite{Zhang.2020}. Furthermore, we considered nature-inspired metaheuristics and evolutionary algorithms, including Tabu with the improved iterated greedy algorithm (TMMIG) \cite{Ding.2015} , coevolutionary quantum genetic algorithm (CQGA) \cite {Deng.2015}, genetic algorithm combined with simulated annealing (HGSA) \cite{Wei.2018}, and a hybrid genetic algorithm with tabu search (GA–TS) \cite{Umam.2022}. We also compared our approach to classic dispatching rules such as FIFO, SPT, SPS, LTWR, SPSP, and LPTN. Finally, we compared against our previous work, PetriRL, in which we used a masked PPO agent to solve the JSSP problem  in the traditional standalone DRL framework \cite{Lassoued.2024}.

\begin{table}[ht]
    \centering
    \footnotesize
    \resizebox{\columnwidth}{!}{%
    \begin{tabular}{ll}
        \toprule
        \multicolumn{2}{c}{\textbf{Heuristics}} \\
        \midrule
        FIFO  & First-In-First-Out. \\
        SPT   & Shortest Processing Time. \\
        SPS   & Shortest Processing Sequence. \\
        LTWR  & Longest Total Work Remaining. \\
        SPSR  & Shortest Processing Sequence Remaining. \\
        LPTN  & Longest Processing Time of Next Operation. \\
        LWT   & Longest Waiting Time. \\
        \addlinespace
        \multicolumn{2}{c}{\textbf{Metaheuristics}} \\
        \midrule
        TMIIG & Tabu Search with Modified Iterated Greedy Algorithm. \\
        CQGA  & Coevolutionary Quantum Genetic Algorithm. \\
        HGSA  & Hybrid Genetic Algorithm with Simulated Annealing. \\
        TSGA  & Tabu Search combined with Genetic Algorithm. \\
        \addlinespace
        \multicolumn{2}{c}{\textbf{Learning-Based}} \\
        \midrule
        GIN   & Graph Isomorphism Network. \\
        GAM   & Gated Attention Model. \\
        DGERD & Disjunctive Graph Embedded Recurrent Decoding\\
              & Transformer. \\
        MPPO  & Maskable Proximal Policy Optimization. \\
              &integrated with Petri net (our previous work). \\
        \bottomrule
    \end{tabular}%
    }
    \caption{Descriptions of all contending algorithms.}
    \label{table:contending-algorithms}
\end{table}

\subsection{Training performance}

Figure~\ref{fig: training_performances} illustrates the training performance of an agent trained on a 20~jobs $\times$ 20~machines Job Shop Scheduling Problem . Four key metrics are reported to assess exploration, exploitation, and learning progress: episode mean reward, combined training loss, entropy loss, and clipping range.

\begin{table*}[ht]
\centering
\begin{tabular}{p{0.8cm}p{0.8cm}p{0.8cm}p{0.8cm}p{0.8cm}p{0.8cm}p{0.8cm}p{0.8cm}p{0.8cm}p{0.8cm}p{0.8cm}p{1cm}p{1cm}}
\toprule
Inst & Size & FIFO & SPT & SPS & LTWR & SPSR & LPTN & LWT & Heur-Avg & Best-Heuristic & Ours & Gap \\
\midrule
ta01 & 15x15  & 1486 & 1454 & 1486 & 1454 & 1486 & 1639 & 1486 & 1761 & \textbf{1454} & \textbf{1454} & 0.0\% \\
ta11 & 20x15  & 1701 & 1771 & 1701 & 1771 & 1671 & 1712 & 1701 & 2083 & 1671 & \textbf{1648} & -1.4\% \\
ta21 & 20x20  & 2089 & 2114 & 2089 & 2114 & 2111 & 2016 & 2089 & 2345 & 2016 & \textbf{1977} & -1.9\% \\
ta31 & 30x15  & 2277 & 2312 & 2277 & 2312 & 2277 & 2260 & 2277 & 2517 & 2260 & \textbf{2219} & -1.8\% \\
ta41 & 30x20  & 2543 & 2661 & 2543 & 2661 & 2543 & 2634 & 2543 & 3080 & \textbf{2543} & \textbf{2543} & 0.0\% \\
ta51 & 50x15  & 3590 & 3564 & 3590 & 3561 & 3590 & 3664 & 3590 & 3792 & 3496 & \textbf{3433} & -1.8\% \\
ta61 & 50x20  & 3690 & 3619 & 3690 & 3619 & 3690 & 3572 & 3690 & 4097 & 3572 & \textbf{3427} & -4.1\% \\
ta71 & 100x20 & 6270 & 6359 & 6270 & 6312 & 6270 & 6282 & 6270 & 6857 & 6248 & \textbf{6190} & -1.6\% \\
\midrule
\textbf{Average} 
& -- 
& 2956 & 2982 & 2956 & 2976 & 2955 & 2972 & 2956 
& 2976 & 2908 & \textbf{2860} & \textbf{-1.6\%} \\
\bottomrule
\end{tabular}
\caption{Performance comparison of heuristic rules and the proposed super-heuristic on Taillard benchmark instances.}
\label{tab: heuristics}
\end{table*}

\begin{table*}[ht]
\centering
\begin{tabular}{p{1cm}p{1cm}p{1cm}p{0.8cm}p{0.8cm}p{0.8cm}p{0.8cm}p{0.8cm}p{0.8cm}p{0.8cm}p{0.8cm}p{0.8cm}}
\toprule
Inst & Size & DGERD & GIN & GAM & TMIIG & CQGA & HGSA & TSGA & MPPO & Best-Heur & Ours \\
\midrule
ta01 & 15x15 & 1711 & 1547 & 1530 & 1486 & 1486 & 1324 & \textbf{1282} & 1436 & 1454 & 1454 \\
ta11 & 20x15 & 1833 & 1775 & 1798 & 2011 & 2044 & 1713 & 1622 & \textbf{1568} & 1671 & 1648 \\
ta21 & 20x20 & 2146 & 2128 & 2086 & 2973 & 2973 & 2331 & 2331 & 2064 & 2016 & \textbf{1977} \\
ta31 & 30x15 & 2383 & 2379 & 2342 & 3161 & 3161 & 2731 & 2730 & 2166 & 2260 & 2219 \\
ta41 & 30x20 & \textbf{2541} & 2604 & 2604 & 4274 & 4274 & 3198 & 3100 & 2576 & 2543 & 2543 \\
ta51 & 50x15 & 3763 & 3394 & 3344 & 6129 & 6129 & 4105 & 4064 & \textbf{3272} & 3496 & 3433 \\
ta61 & 50x20 & 3633 & 3594 & 3534 & 6397 & 6397 & 5536 & 5502 & 3505 & 3572 & \textbf{3427} \\
ta71 & 100x20 & 6321 & 6098 & 6027 & 8077 & 8077 & 5964 & \textbf{5962} & 6366 & 6248 & 6190 \\
\midrule
\textbf{Average} 
& -- & 3041 & 2940 & 2908 & 4314 & 4318 & 3363 & 3324 & 2869 & 2908 & \textbf{2860} \\
\bottomrule
\end{tabular}
\caption{Comparison of the proposed super-heuristic with representative approaches from the literature on Taillard benchmarks.}
\label{tab:taillard_superheuristic}
\end{table*}

\begin{figure}[ht]
\centering
\includegraphics[width=\linewidth]{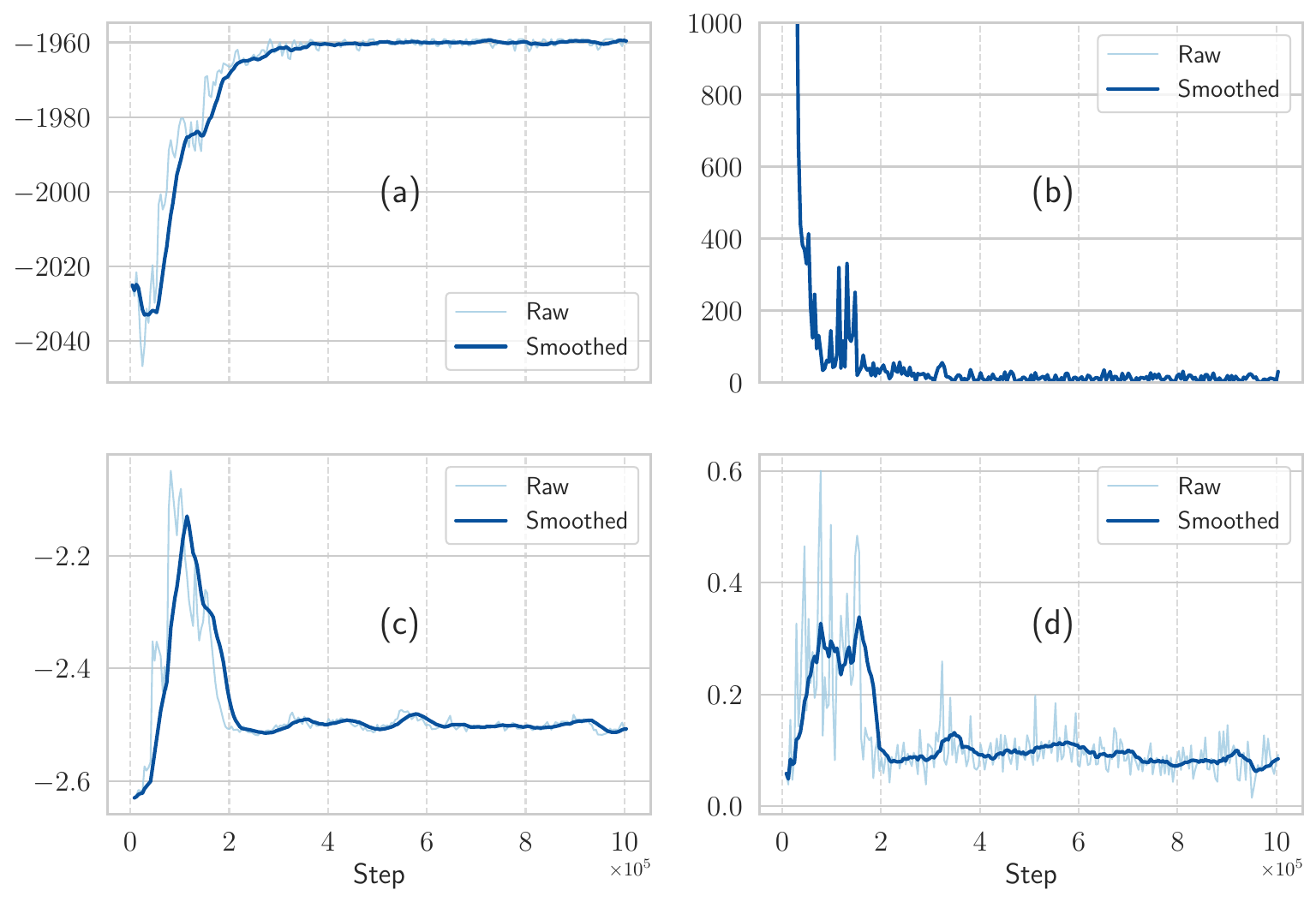}
\caption{Agent training performances on a 20 jobs x 20 machines instance . (\textbf{a})  the episode mean reward  , (\textbf{b}) the combined training loss,(\textbf{c}) the entropy loss (\textbf{d}) the clipping range. The agent is trained for \textbf{1e6} steps with 5-step commitment using the maskable PPO algorithm.}
\label{fig: training_performances}
\end{figure}

Subplot (a) shows a steady increase in the collected reward, which directly reflects the makespan. In an effort to reduce the incurred penalty, the agent works to minimize the makespan while respecting the constraints imposed by the environment. Subplot (b) depicts the total loss, combining the losses from the policy network and the critic. This confirms that the critic is becoming better at correctly predicting the values needed to calculate the advantage, a crucial component of the policy network's gradient.

Finally, subplots (c) and (d) show exploration behavior during the first 200,000 steps, characterized by high policy entropy. This led the agent to clip more (approximately 30\%) to maintain training stability, as demonstrated in subplot (d). Both subplots indicate a transition toward exploitation, with decreased entropy and fewer action updates being clipped.

\subsection{Results analysis}

In line with the claims from the literature discussed in sub-section~\ref{subsec: Policy Search over Heuristics}, we empirically confirm that the learned HH outperforms the best single heuristic on the scheduling task, as evaluated on the Taillard benchmark instances. Table~\ref{tab: heuristics} presents the makespan results across problem instances ranging from $15 \times 15$ to $100 \times 20$ jobs and machines. Across all instance sizes, the HH performed similarly to or better than the static heuristics. The improvement ranges from 1.4\% to 4.1\%, with an average improvement of 1.6\% relative to the best heuristic and 4\% relative to the average performance of the heuristics.

This improvement can be explained by the state-dependent switching behavior discussed in sub-section~\ref{subsec: State-Dependent Switching}. Unlike static low-level heuristics, HH can adapt dynamically to the current system state, thereby improving condition handling on the shop floor.

After comparing the performance of the HH algorithm with traditional heuristic rules and demonstrating that it consistently matches or outperforms the best heuristic for any given instance of the Taillard benchmark, we now turn to benchmarking against approaches from the literature listed in the Table~\ref {table:contending-algorithms}. 

On average across all instance sizes, the HH approach outperforms all benchmarked algorithms, achieving the lowest average makespan of 2860 steps. While its dominance over fixed heuristics is clear, comparisons with metaheuristics and learning-based approaches are more competitive, as each method may excel on specific instance sizes but not consistently across all instances. Nevertheless, the HH retains a distinct advantage in explainability. Unlike neural networks, which often function as black-box mappings from states to actions, HH makes decisions using a set of interpretable dispatching rules. For instance, if the agent allocates an operation from job two under the MTWR rule, the decision can be directly traced to the job with the most remaining work. This transparency provides a more interpretable decision-making process while sustaining strong performance across diverse scheduling instances.

\subsection{Ablation}
\label{sec: ablation}

In the ablation study, we first qualitatively analyze the impact of employing a hyper-heuristic, rather than acting directly in the environment's action space, on training performance. This is followed by a quantitative assessment of the impact of the number of commitment steps on the resulting makespan. Additionally, we evaluate the effects of different action-selection strategies during inference, considering both a greedy and a sampling-based approach.

\subsubsection{Commitment steps}

\begin{figure}[ht]
\centering
\includegraphics[width=\linewidth]{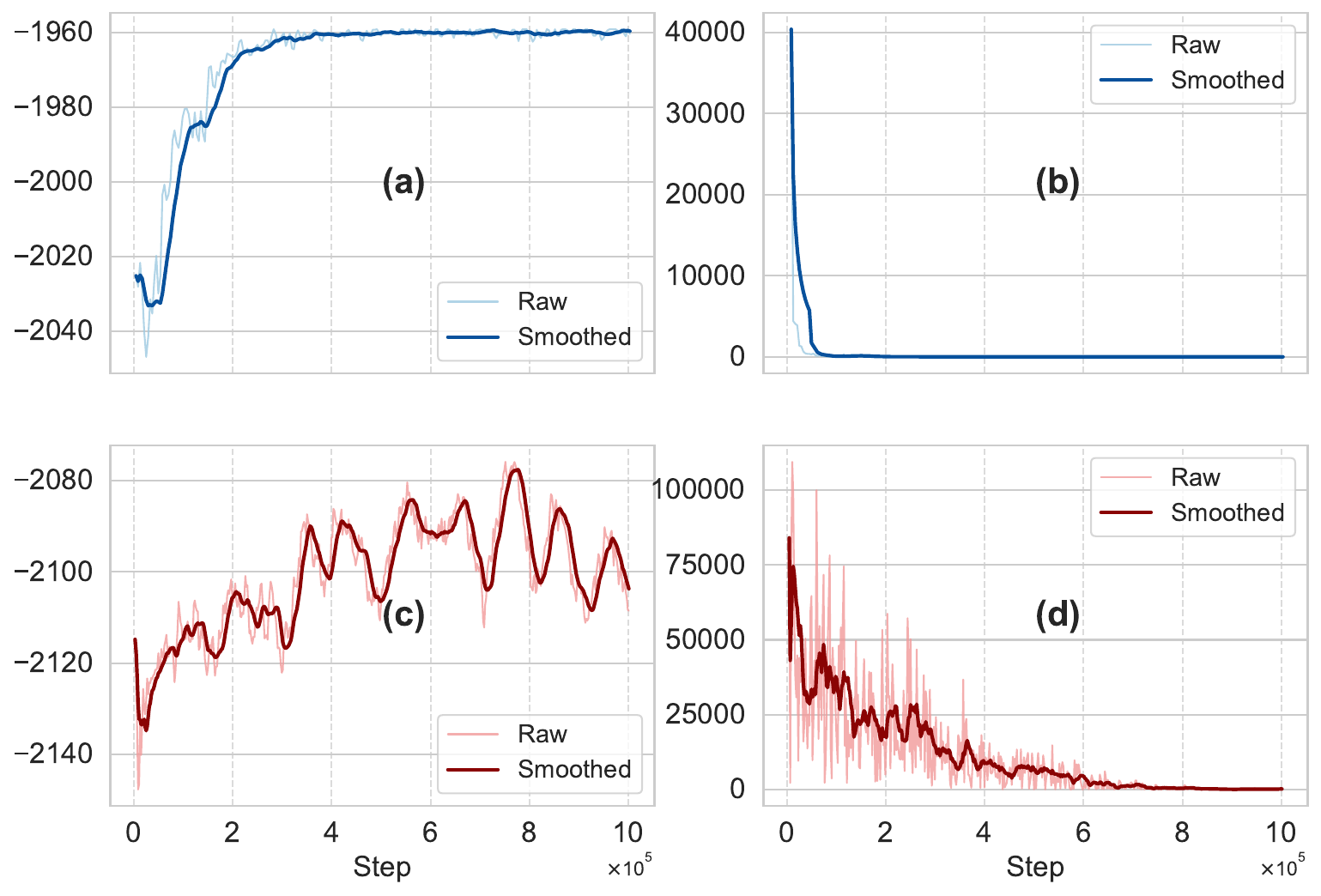}
\caption{Comparison of two control strategies applied to the same environment and problem instance (20 jobs x 20 machines): a hyperheuristic agent selecting among a set of heuristic rules (blue) and a reinforcement learning agent acting directly on the environment's action space (red). Subfigures show: (a) rewards collected by the hyperheuristic agent, (b) its associated training loss function, (c) rewards collected by the direct-action RL agent, and (d) its corresponding training loss function.}
\label{fig:stability}
\end{figure}  

In Figure~\ref{fig:stability}, we compare solving a 20~jobs~$\times$~20~machines JSSP using two cases. In the first case, a PPO agent acts directly on the environment's action space, whereas in the second case, an HH agent operates on a heuristic action space with a 5-step commitment. Qualitatively, both the collected reward function and the training loss are noticeably smoother in this case, indicating more stable training. The commitment mechanism improves credit assignment by effectively shortening the original non-heuristic trajectory length by a factor equal to the number of commitment steps.

Quantitatively, the HH agent achieves a maximum reward of approximately $-1960$ (corresponding to a makespan of 1960) and stabilizes after roughly $4\times10^{4}$ steps. In contrast, the direct RL agent peaks at approximately -2080 and exhibits substantially more unstable performance.

\begin{table}[htbp]
\centering
\caption{effect of commitment horizon on performance}
\label{tab:performance_comparison}
\begin{tabular}{lccc}
\toprule
Instance size & 1-step & 5-steps & 1000-steps \\
\midrule
\addlinespace[2pt]
15$\times$15   & 1486 & 1454 & 1454 \\
20$\times$15   & 1671 & 1648 & 1671 \\
20$\times$20   & 1979 & 1977 & 2016 \\
30$\times$15   & 2217 & 2219 & 2260 \\
30$\times$20   & 2554 & 2530 & 2543 \\
50$\times$15   & 3528 & 3433 & 3496 \\
50$\times$20   & 3718 & 3417 & 3572 \\
100$\times$20  & 6170 & 6190 & 6248 \\
\addlinespace[2pt]
\midrule
\addlinespace[2pt]
Average        & 2915 & \textbf{2860} & 2908 \\
\addlinespace[2pt]
\bottomrule
\end{tabular}
\end{table}

We also evaluated the impact of commitment length by comparing three configurations: 1-step, 5-step, and 1000-step commitment. The 1-step commitment represents the lower bound of our approach, equivalent to continuous rule switching at every timestep. The 5-step commitment is our default configuration used throughout this paper. The 1000-step commitment represents the upper bound, effectively degenerating to a fixed single-heuristic policy, since all benchmark instances have trajectories requiring fewer than 1000 action selections, forcing the agent to commit to one rule for the entire episode.

The results demonstrate that the 5-step commitment achieves the best performance, outperforming both the single best static heuristic and the continuous switching baseline. Consistent with our previous findings that intermediate commitment lengths provide superior training stability, this stability translated into a lower average makespan than continuous switching, while maintaining the flexibility to adapt heuristic selection throughout the scheduling process. In contrast, a 1000-step commitment eliminates this adaptability, reducing performance to that of a fixed heuristic policy.

\subsubsection{Action selection}

Finally, we tested the effect of two action selection methods during inference: greedy deterministic selection and sampling. In the greedy deterministic method, the algorithm selects the action with the highest softmax probability, i.e., the action with the highest logit value, using the argmax function. On the other hand, in the sampling method, the action is chosen by sampling from the categorical probability distribution produced by the softmax function, introducing some randomness into the decision process. We used the deterministic method throughout the study.

The results show that the makespan using sampling or deterministic selection yielded similar results, and we can attribute this phenomenon to two main reasons. First, in multinomial sampling, each action \( a_i \) is selected according to the probability \( p_i \) as follows: \( P(\hat{a} = a_i) = p_i \) for \( i = 1, 2, \dots, n \). If the agent is trained sufficiently, the probability distribution of a given action becomes so dominant that even with sampling, the agent almost always selects that action. 

Additionally, some heuristic rules yield the same decision under a given state, making them equivalent. This can be seen when analyzing the probability distribution in a given state. In this case, two actions are co-dominant with similar probabilities. Despite the fact that the RL agent alternates between selecting the two actions, the final makespan remains the same. In other words, the two different heuristics, under the given constraints, select the same action (e.g., selecting the same job for a machine), especially in highly constrained problems like JSSP, the valid action selection could be limited.

Exploration is a cornerstone of the RL framework. Like in inference, environments with discrete action spaces often rely on multinomial sampling to explore the solution space during training. However, multinomial sampling can become limiting if one action's probability dominates the distribution early, thereby hindering the agent from exploring other parts of the solution space. Further studying this phenomenon and finding alternatives could enhance the exploration phase in RL.

\section{Conclusion}
\label{sec: Conclusion}

In this paper, we addressed the Job Shop Scheduling Problem using a deep policy-based reinforcement learning hyper-heuristic (DRL-HH) framework. To this end, we extended the classical hyper-heuristic (HH) framework by two key mechanisms: action masking and action commitment.

Invalid actions are pre-filtered before reaching the low-level heuristics (LLHs) for decision-making. This is achieved through the Petri net's dynamic action masking capability, which disables transitions according to guard-function rules. Unlike traditional HH approaches that assess action validity post hoc via move-acceptance strategies, often penalizing LLHs for infeasible decisions, our method ensures that LLHs are evaluated exclusively on valid actions. This separation enables the high-level policy to learn heuristic selection independently of hard environmental constraints, thereby allowing a more objective and unbiased evaluation of the different LLHs.

We evaluated the proposed framework against a broad range of competing methods, including classical heuristics, metaheuristics, and recent deep reinforcement learning approaches such as Graph Isomorphism Networks, Gated Attention Models, and Transformer-based architectures. Across multiple Taillard benchmark instances, our method consistently achieved a superior average makespan.

In addition, we introduced action commitment, inspired by temporally extended actions in the Options framework (macro-actions). Per-step switching tends to exacerbate credit assignment issues, while rigidly committing to a single heuristic for an entire episode limits adaptability and often leads to suboptimal performance. As an alternative, we define a commitment horizon as a tunable hyperparameter that controls the duration for which a selected heuristic is applied. 

Ablation studies show that intermediate commitment lengths provide the best trade-off. Specifically, they qualitatively improve training stability and quantitatively achieve a lower makespan than both per-step heuristic switching and full-episode commitment.

Finally, we investigated different action selection strategies at the policy level, comparing deterministic greedy selection with stochastic sampling. The results were essentially identical, which we attribute to two combined effects. First, early dominance of specific action probabilities in the softmax output causes multinomial sampling to collapse to near-deterministic behavior. The second structural constraint imposed by the Petri net leads multiple LLHs to produce identical low-level actions under the same state. 

Early probability dominance can limit exploration during training, particularly in the initial learning phase. Addressing this issue represents a promising direction for future work, potentially enabling richer exploration and the discovery of more diverse scheduling strategies.

Overall, this work demonstrates that structurally informed action filtering and temporal abstraction can enhance reinforcement learning–based hyper-heuristics. 

\bibliographystyle{elsarticle-num} 
\bibliography{references}

\end{document}